\documentclass[11pt]{article}

\usepackage[final]{acl}

\usepackage{times}
\usepackage{latexsym}

\usepackage[T1]{fontenc}

\usepackage[utf8]{inputenc}

\usepackage{microtype}

\usepackage{inconsolata}

\usepackage{graphicx}
\graphicspath{ {figures} }

\usepackage{xurl}
\usepackage{hyperref}
\usepackage{url}
\usepackage{booktabs}
\usepackage{lineno}
\usepackage[listings,skins,breakable]{tcolorbox}
\usepackage{tabularx}
\usepackage[table]{xcolor}
\usepackage{caption}
\usepackage{multirow}
\usepackage{amsfonts}
\usepackage{amsmath}
\usepackage{array}
\usepackage{amssymb}
\usepackage{pifont}
\usepackage{multirow}
\usepackage{ragged2e}
\usepackage{siunitx}
\newcommand{\cmark}{\ding{51}}%
\newcommand{\xmark}{\ding{55}}%
\newcommand{\mse}{\text{MSE}}
\newcommand{\se}{\text{SE}}

\definecolor{SuccessColor}{HTML}{1ABC9C}
\definecolor{FailureColor}{HTML}{E74C3C}
\definecolor{HeaderGray}{gray}{0.95}
\newcommand{\Success}{\textcolor{SuccessColor}{\cmark}}
\newcommand{\Failure}{\textcolor{FailureColor}{\xmark}}
\newcommand{\Attr}[3]{\texttt{#1}: #2 $\rightarrow$ \textbf{#3}}
\newcommand{\Note}[1]{\textit{\small #1}}
\newcommand{\Target}[1]{\textbf{Target:} \textit{#1}}

\title{LingGen: Scalable Multi-Attribute Linguistic Control via Power-Law Masking}

\author{Mohamed Elgaar \and Hadi Amiri \\
  University of Massachusetts Lowell\\
  \texttt{\{melgaar,hadi\}@cs.uml.edu}}

\begin{document}
\maketitle
\begin{abstract}
We present LingGen, a controlled text generation model that allows fine-grained control over a large number of real-valued linguistic attributes. It encodes target attribute values with a dedicated linguistic attribute encoder and conditions the language model by injecting the resulting representation into the language model using the beginning-of-sequence (BOS) embeddings. To improve robustness when controlling different attribute subsets, we introduce P-MASKING, which samples per-example attribute masking rates from a truncated Pareto distribution during training. Across 1--40 control attributes, LingGen achieves the lowest average control error among evaluated methods, while remaining efficient at inference and receiving the highest fluency scores in human evaluation. Ablations show that Pareto-sampled masking and BOS-based injection are effective choices compared to alternative masking and integration variants.\footnote{Code \& data: \url{https://github.com/CLU-UML/LingGen}}
\end{abstract}

\section{Introduction}
Controlled text generation (CTG) aims to produce text that satisfies user-specified constraints, with applications in education, accessibility, and personalized communication~\citep{Dathathri2020Plug,zhang2023survey,prabhumoye2020exploring}. While existing approaches can control coarse properties (e.g., sentiment), fine-grained linguistic control remains difficult~\citep{liu-etal-2023-composable}, especially when (i) target linguistic attributes are real-valued, (ii) the number of controlled attributes varies at inference time, and (iii) the attribute set is large enough such that interactions and trade-offs become unavoidable. In this work, we study scalable linguistic control with \(k=40\) expert-crafted linguistic complexity attributes spanning surface counts, syntactic structure, and psycholinguistic complexity (Appendix~\ref{sec:index_list}).

Existing CTG methods can steer high-level attributes such as  sentiment or topic, but they are less effective for finer-grained linguistic attributes and suffer from inefficiencies when many attributes must be handled jointly~\citep{li-etal-2018-delete,liu-etal-2023-composable}. In parallel, recent work shows that combining denoising objectives with causal language modeling can improve robustness~\citep{raffel2020exploring, tayul2, zengglm, wettig-etal-2023-mask}. Building on these insights, we propose LingGen--a controlled generation model that supports fine-grained control over a large number of real-valued linguistic attributes. 

LingGen conditions a base language model on target attribute values using a dedicated attribute encoder and a lightweight integration mechanism: it injects the encoded attribute representation into the beginning-of-sequence (BOS) embedding, allowing self-attention to propagate conditioning information without modifying every token embedding. To generalize control to arbitrary subsets of attributes (from 1 to $k$), we introduce P-MASKING, which stochastically hides attributes during training by sampling a masking rate from a truncated Pareto distribution (a bounded power law distribution). This sampling scheme emphasizes near-complete attribute sets while preserving a tail of sparse configurations to improve robustness when the number of controlled attributes varies at test time.\looseness-1

\begin{table*}[t]
    \centering
    \small
    \begin{tabular}{lcccc}
    \toprule
    \textbf{Capability} & \textbf{LingGen} & \textbf{Fine-tuning} & \textbf{Inference-time} & \textbf{LLM Prompting} \\ & \small{(Ours)} & \small{(MCTune, PTG)} & \small{(PPLM, Guided Decoding)} & \small{(Llama 3.1)} \\
    \midrule
    Fine-grained control & \textcolor{green!60!black}{\cmark} & \textcolor{green!60!black}{\cmark} & \textcolor{green!60!black}{\cmark} & \textcolor{red!80!black}{\sffamily \xmark} \\
    Maintains high fluency & \textcolor{green!60!black}{\cmark} & \textcolor{green!60!black}{\cmark} & \textcolor{red!80!black}{\sffamily \xmark} & \textcolor{green!60!black}{\cmark} \\
    Efficient at inference & \textcolor{green!60!black}{\cmark} & \textcolor{green!60!black}{\cmark} & \textcolor{red!80!black}{\sffamily \xmark} & \textcolor{red!80!black}{\sffamily \xmark} \\
    Robust \& scalable attribute control$^{\dagger}$ & \textcolor{green!60!black}{\cmark} & \textcolor{red!80!black}{\sffamily \xmark} & \textcolor{red!80!black}{\sffamily \xmark} & \textcolor{red!80!black}{\sffamily \xmark} \\
    \bottomrule
    \end{tabular}
    \vspace{-2mm}
    \caption{Comparison of LingGen with other CTG approaches. $^{\dagger}$Maintains effective control over any subset of attributes, including high-count subsets (>20).}
    \label{tab:comparison}
    \vspace{-4mm}
\end{table*}

Our contributions are as follows: 
(\textbf{1}) we propose LingGen, a CTG model for real-valued, fine-grained linguistic control that scales to 40 attributes and supports controlling any subset size (1--40) at inference time;
(\textbf{2}) we introduce P-MASKING, a Pareto-sampled, per-example attribute masking scheme that improves robustness across attribute subset sizes;
(\textbf{3}) we provide empirical analysis of attribute interactions and benchmark LingGen against fine-tuning, inference-time control, and prompting baselines, and highlight the trade-offs between {\em control} accuracy, {\em fluency}, and {\em efficiency}.

\section{Background}
CTG has increasingly focused on methods to regulate multiple attributes simultaneously, such as sentiment, tense, formality, or specific keywords~\citep{shen2017style}. However, traditional models often lack the flexibility to adapt to new configurations, leading to inefficiencies and quality degradation when handling multiple controls, especially with finer-grained linguistic attributes~\citep{li-etal-2018-delete,liu-etal-2023-composable}.

Recent advancements have explored compositional text control in latent space by leveraging compact, differentiable representations. Techniques based on ordinary differential equations and latent space samplers can efficiently compose multiple control operations while reducing computational overhead and maintaining text quality~\citep{liu-etal-2023-composable,ding-etal-2023-maclasa}. These methods align with the growing interest in developing models that adapt to diverse control inputs across various domains~\citep{yang-etal-2023-tailor}.

In parallel, research into Masked Language Models (MLMs) has highlighted the effectiveness of masking strategies~\citep{devlin-etal-2019-bert}. Our method draws on prior work such as PMI-Masking~\citep{levine2020pmi} and infilling objectives in UL2 and GLM-130B~\citep{tayul2,zengglm,levine2020pmi}, but targets a different problem: robust conditioning on a variable-size set of linguistic attribute controls. While traditional MLMs use a fixed 15\% masking rate~\citep{devlin-etal-2019-bert}, recent work has shown benefits from higher rates---up to 40\% or 80\%---in some settings~\citep{wettig-etal-2023-mask}. Building on these insights, we design P-MASKING, which masks linguistic control attributes during training according to a power law distribution~\citep{clauset2009power}. This distribution exposes the model to both dense and sparse attribute sets.

CTG enables text creation tailored to specific requirements, with recent works exploring various approaches such as sentiment manipulation via fine-grained control codes~\citep{shi2024lifi}, tunable biases for factual consistency~\citep{liu-etal-2023-bolt}, and prefix-adaptive decoding for style control~\citep{pei-etal-2023-preadd}.
However, these methods primarily focus on high-level properties rather than low-level linguistic attributes.
Models incorporating denoising objectives in pretraining, such as UL2 and GLM, have demonstrated enhanced capabilities in diverse linguistic tasks~\citep{tayul2, zengglm,chowdhery2023palm, roberts2023scaling}.
Unlike existing approaches that use instruction-tuning~\citep{nguyen-etal-2024-multi}, prompt-tuning~\citep{bandel-etal-2022-quality,alhafni-etal-2024-personalized,yang-etal-2023-tailor}, concatenation~\citep{huang-etal-2023-extensible}, or simple fusion~\citep{liu-etal-2023-composable}, LingGen introduces a dedicated attribute embedding network and selective masking to control a variable number of attributes while preserving base LLM capabilities.
In addition, although multi-attribute controlled {\em paraphrase} generation has been studied~\citep{elgaar-amiri-2025-linguistically}, LingGen targets {\em free text} generation with robust control over {\em variable-size} attribute subsets.
Table~\ref{tab:comparison} provides a summary of how LingGen compares to existing approaches.

\begin{figure*}[t]
    \centering
    \includegraphics[width=0.9\linewidth]{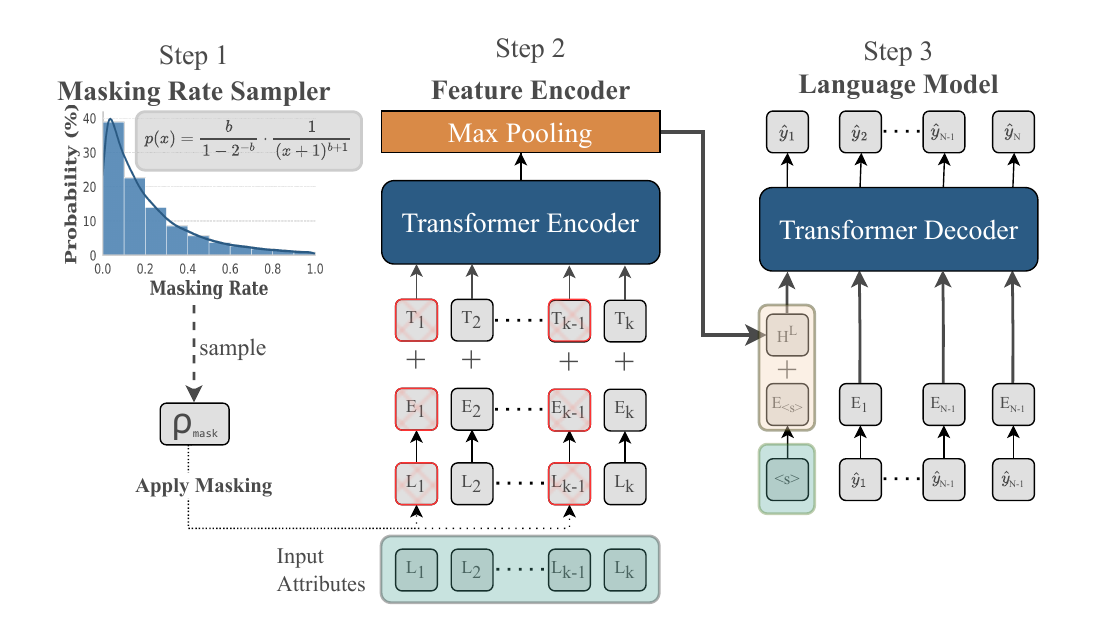}
    \vspace{-2mm}
    \caption{
        Overview of the LingGen architecture for controlled text generation. 
        1) \textbf{Masking Rate Sampler (Training):} Implements P-MASKING by sampling attribute masking rates ($\rho_{mask}$) from a Pareto distribution per sample.
        2) \textbf{Attribute Encoder:} Encodes linguistic attributes ($L_1..L_K$) into a combined representation using embeddings and attribute-specific token types ($T_1..T_K$).
        3) \textbf{Language Model:} A Transformer Decoder generates text ($\hat{y}_1..\hat{y}_n$) conditioned on the attribute representation, which is injected into the BOS token (\texttt{<s>}) embedding to steer generation.
        \looseness-1
    }
    \label{fig:arch}
    \vspace{-10pt}
\end{figure*}

\begin{table*}[t]
    \centering
    \begin{tabular}{lrr>{\raggedleft\arraybackslash}p{1.8cm}>{\raggedleft\arraybackslash}p{1.1cm}}
    \toprule
    \textbf{Method} & \textbf{MSE $\downarrow$} & \textbf{PPL $\downarrow$} & \multicolumn{2}{c}{\textbf{Time/token (ms) $\downarrow$}} \\
    \midrule
    \hspace{-0.15cm}\textbf{No Control} & & & & \\
    Reference & 0.00 & 35.4 & - & - \\
    Vanilla OPT-350M~\footnotesize\citep{zhang2022opt} & 2.00 & 12.7 & 25 & (1$\times$) \\
    \midrule
    \hspace{-0.15cm}\textbf{Inference-time Control} & & & & \\
    PPLM~\footnotesize\citep{dathathriplug} & 5.99 & \underline{10.7} & 1515 & (61$\times$) \\
    Guided Decoding~\footnotesize\citep{glandorf-etal-2025-grammar} & 3.74 & 20.5 & 112 & (5$\times$) \\
    COLD~\footnotesize\citep{qin2022cold} & 3.99 & 11.3 & 3846 & (155$\times$) \\
    Mix\&Match~\footnotesize\citep{mireshghallah-etal-2022-mix} & 1.58 & 165.0 & 5882 & (237$\times$) \\
    BOLT~\footnotesize\citep{liu-etal-2023-bolt} & 2.08 & 12.0 & 114 & (5$\times$) \\
    PiLM~\footnotesize\citep{yang-etal-2024-plug} & 2.12 & 17.7 & 2102 & (84$\times$) \\
    Llama 3.1~\footnotesize\citep{dubey2024llama} & 3.11 & 14.5 & 162 & (7$\times$) \\

    \midrule
    \hspace{-0.15cm}\textbf{Fine-tuned LLM} & & & & \\
    MCTune {\tiny Llama 2-7B}~\footnotesize\citep{nguyen-etal-2024-multi} & 2.44 & \textbf{5.8} & 68 & (3$\times$) \\
    MCTune {\tiny OPT-350M}~\footnotesize\citep{nguyen-etal-2024-multi} & 11.46 & 16.4 & 46 & (2$\times$) \\
    PTG~\footnotesize\citep{alhafni-etal-2024-personalized} & \underline{1.71} & 21.5 & 76 & (3$\times$) \\
    LingGen (ours) & \textbf{1.08} & 15.4 & \textbf{25} & \textbf{(1$\times$)} \\
    \bottomrule
    \end{tabular}
    \caption{Comparison of model performance across different methods. The table presents the average MSE across varying numbers of attributes (1, 5, 10, 20, 40), perplexity (PPL), and time per token.
    }
    \label{tab:main_results}
    \vspace{-5pt}
\end{table*}
\section{Linguistic Generation with LingGen}

Given a set of desired linguistic attributes, $\pmb{a} = \{L_1, \dots, L_k\}$, where each $L_i$ represents a specific linguistic attribute (e.g., sentence length, number of unique sophisticated words), the task is to generate text that exhibits those attributes. In our setup, target attribute values are obtained by running the deterministic attribute extractor on held-out reference texts and z-normalizing each attribute using training-set mean and standard deviation; at test time we sample a reference text, randomly select the requested number of attributes (e.g., 1/5/10/20/40), and use their normalized values as the control targets. We use 40 attributes (detailed in Appendix~\ref{sec:index_list}) representing diverse dimensions of linguistic style with low mean Pearson correlation (0.29), indicating minimal redundancy. Fine-grained stylistic control often requires simultaneous manipulation of complex features; for instance, generating accessible materials for individuals with aphasia requires control over word frequency, syntactic complexity, and lexical density beyond simple readability scores. Our P-MASKING strategy is designed to improve control across a variable and large number of attributes (from 1 to $k$).
Let $Y$ be the space of possible generated texts. Our goal is to find a model $G$ that takes the desired attributes $\pmb{a}$ as input and generates a text $\pmb{y} = G(\pmb{a})$ that minimizes a loss function $L(V(\pmb{y}), \pmb{a})$, where  $V: Y \rightarrow \mathbb{R}^k$ is a function that extracts a fixed-size vector representation of the attributes present in a given text~\citep{hu2017toward}.
This can be expressed as finding $\pmb{y} = \arg\min_{\pmb{y} \in Y} L(V(\pmb{y}), \pmb{a})$.
Note that there can be multiple solutions $\pmb{y}$ that minimize this loss.
For example, if $\pmb{a}$ specifies a sentence of length 10, there are many possible sentences of length 10 that could be generated.
However, as the number of attributes in $\pmb{a}$ increases and the granularity of these attributes becomes finer (e.g., specifying not just sentence length but also specific keywords, syntactic structure, and sentiment),
the set of possible solutions shrinks. In the extreme case, with a sufficiently large and specific set of attributes,
there may be only one or a very small number of sentences $\pmb{y}$ that satisfy all the constraints~\citep{holtzman2019curious}.

We train the model using cross-entropy loss on the predicted token sequence, conditioned on the input attributes. Cross-entropy loss is particularly useful because it aligns with the model's training objective of predicting the next word in a sequence, thus reducing the discrepancy between training and test conditions. This helps mitigate the accumulation of errors during sequence generation, as the model learns to generate text that is both fluent and coherent while conforming to the desired attributes~\citep{ranzato2016sequence,bengio2000neural}. Training on a large and diverse dataset with a wide variety of attribute combinations allows the model to learn the underlying relationship between attributes and text,
enabling it to generate text that is both fluent and coherent while conforming to the desired attributes~\citep{radford2019language}. The attribute values themselves are computed using explicit linguistic analysis algorithms from prior work~\citep{Lu_2010,lu2012relationship,lee-lee-2023-lftk}.
These algorithms provide the function $V(\pmb{y})$ that maps generated text $\pmb{y}$ to a vector of attribute values in $\mathbb{R}^k$. These same algorithms are used during evaluation to provide reliable, deterministic measurement of attribute control error, ensuring that our reported MSE values reflect true deviations from the target attributes rather than relying on approximations.

LingGen consists of a Masking Rate Sampler, an Attribute Encoder, and an LLM (Figure~\ref{fig:arch}), working together to allow for flexible control over a variable number of attributes.

\subsection{Attribute Integration}
The $k$ linguistic attributes $\pmb{a} = \{L_1, \dots, L_k\}$ are processed through an Attribute Encoder architecture. First, the encoder employs a linear embedding layer that maps each scalar attribute value $L_i$ to a $d_{attr}$-dimensional embedding space through the transformation $E_i = W_{emb}L_i + b_{emb}$ where $W_{emb} \in \mathbb{R}^{d_{attr} \times 1}$ and $b_{emb} \in \mathbb{R}^{d_{attr}}$. These attribute embeddings are then combined with learned token-type embeddings $T_i$ that indicate the attribute identity (e.g., sentence length vs.\ lexical sophistication), which is particularly important when only a subset of attributes is provided as input.

The combined embeddings are processed by a transformer encoder consisting of 2 layers with 5 attention heads, which captures the interactions between different attributes. The encoder's output undergoes max-pooling across the sequence dimension to obtain a fixed-size representation, which is then mapped to the model's hidden dimension through a final projection layer $W_{proj} \in \mathbb{R}^{d_{attr} \times d_{LLM}}$.

A key design choice is to add the resulting attribute representation only to the beginning-of-sequence (BOS) token embedding. This allows the conditioning signal to influence generation through self-attention while avoiding the distribution shift that can arise when adding conditioning to every token embedding.


\subsection{P-MASKING: A Sample-based Attribute Masking Strategy}
\label{sec:pmasking}

During training, we employ P-MASKING, a strategy that selects the number of linguistic control attributes to mask for each sample. Intuitively, varying which attributes are visible encourages the model to rely on the provided controls (rather than a fixed attribute configuration) and improves robustness when the requested subset of attributes changes at inference time.

P-MASKING samples masking rates from a truncated Pareto distribution~\citep{burroughs2001upper}. For each training sample, a masking proportion $\rho_{\text{mask}}$ is sampled, determining the proportion of attributes to mask. This allows the model to learn to control any number of attributes. Notably, this strategy introduces only a single hyperparameter ($b$) keeping tuning complexity minimal. The probability density function for masking proportion $\rho_{\text{mask}}$ is given by:
\begin{equation}
p(x) = \frac{b}{1-2^{-b}} \frac{1}{(x+1)^{b+1}}
\end{equation}
for $0 \le x \le 1$, where $b$ is a shape parameter. We sweep $b \in \{1, 2, 5, 10, 20\}$ and measure the resulting masking-rate histogram (Appendix~\ref{app:pareto_shape}). As shown in Figure~\ref{fig:pareto_shape}, the selected value $b=5$ concentrates probability mass on near-complete attribute sets (0--30\% masking) while still assigning some probability to sparse configurations (50--99\% masking). The sampled masking proportion $\rho_{\text{mask}}$ determines how many attributes are masked. Masked attributes are excluded from the self-attention and from the output representation pooling.

This sample-based strategy introduces per-example variation in the visible attribute set. The power law distribution yields frequent low masking rates (dense conditioning) while still sampling sparse cases that are necessary for subset robustness.

We aim for the model to perform well on any subset of attributes. However, using a fixed masking rate would train the model primarily on subsets of a certain size (e.g., $k \times (1-\rho_{\text{mask}})$ visible attributes), potentially limiting its ability to handle the full set of $k$ attributes together or single attributes in isolation.
Therefore, P-MASKING samples the number of masked attributes from a power law distribution: it places high probability on low masking (nearly complete visible sets) and a low-probability tail on high masking (sparse visible sets). Dense cases support learning precise multi-attribute control, while sparse cases train the model to remain functional when few attributes are provided.

\section{Experiments}
\subsection{Baselines}
\label{sec:baselines}
We compare against the following baselines:
\paragraph{Vanilla OPT-350M}~\citep{zhang2022opt} generates text by sampling from the LLM without attribute conditioning, serving as a control to verify attribute learning.
\paragraph{Reference} uses the reference sentence as output, providing an upper performance bound by assuming perfect reproduction of the reference text.
\paragraph{Mix\&Match}~\citep{mireshghallah-etal-2022-mix} treats controllable generation as sampling from an energy-based model, using MLM scores and an attribute discriminator.
\paragraph{Plug and Play Language Models (PPLM)}~\citep{dathathriplug} combines a pre-trained LLM with attribute classifiers for attribute control without LLM retraining.
\paragraph{Guided Decoding}~\citep{glandorf-etal-2025-grammar} is an adaptation of FUDGE~\citep{yang-klein-2021-fudge} that steers generation by modifying output probabilities to satisfy constraints.
\paragraph{Llama 3.1-70B}~\citep{dubey2024llama} is an LLM prompting baseline using the instruction-tuned Llama 3.1 (70B) chat model.
\paragraph{Biases Over Logits (BOLT)}~\citep{liu-etal-2023-bolt} directly modifies LLM output logits using learned biases, tuned to minimize attribute and perplexity losses.
\paragraph{COLD Decoding}~\citep{qin2022cold} frames constrained generation as an energy minimization problem, employing gradient-based sampling for constraint adherence.
\paragraph{Multi-Control Tuning (MCTune)}~\citep{nguyen-etal-2024-multi} uses instruction tuning for linguistic attribute control, incorporating attribute definitions into a meta prompt and appending target values to input prompts.
\paragraph{Personalized Text Generation with Fine-Grained Linguistic Control (PTG)}~\citep{alhafni-etal-2024-personalized} controls linguistic attributes via a learned representation prepended to the input text.
\paragraph{Plug-in Language Model (PiLM)}~\citep{yang-etal-2024-plug} adjusts LLM latent states for control using black-box attribute discriminators.

\begin{table}[t]
    \centering
    \begin{tabular}{lr}
    \toprule
    \textbf{Method} & \textbf{Fluency $\uparrow$} \\
    \midrule
    BOLT & 3.73 \\
    PiLM & 3.80 \\
    PTG & 4.20 \\
    LingGen (ours) & \textbf{4.60} \\
    \bottomrule
    \end{tabular}
    \caption{Human evaluation of fluency for each method. The scores range from 1 to 5.}
    \label{tab:human_eval}
\end{table}

\begin{table*}[t]
    \centering
    \begin{tabular}{lrrrr}
    \toprule
    \textbf{Method} & \textbf{\# Sophisticated} & \textbf{\# Total} & \textbf{\# Lexical} & \textbf{Unique Word Ratio} \\
    \midrule
    Vanilla OPT-350M & 3.56 & 14.08 & 6.89 & 11\% \\
    COLD & 3.75 & 19.31 & 9.52 & 17\% \\
    Mix\&Match & 3.83 & 10.67 & 6.55 & 10\% \\
    PPLM & 7.06 & 50.64 & 23.56 & 30\% \\
    PiLM & 3.61 & 18.65 & 9.62 & 14\% \\
    Guided Decoding & 5.91 & 40.63 & 17.93 & 14\% \\
    BOLT & 3.58 & 25.21 & 10.71 & 14\% \\
    MCTune {\tiny OPT-350M} & 5.84 & 17.04 & 10.04 & 12\% \\
    MCTune {\tiny Llama 2-7B} & 5.16 & 16.24 & 9.19 & 11\% \\
    Llama 3.1-70B & 4.30 & 10.43 & 6.04 & 8\% \\
    PTG & 3.10 & 8.44 & 5.20 & 9\% \\
    LingGen & \textbf{1.28} & \textbf{1.76} & \textbf{1.50} & \textbf{6\%} \\
    \bottomrule
    \end{tabular}
    \caption{Mean Absolute Error (MAE) for selected linguistic attributes across different models. The values represent the error between the attributes of the generated text and the target attributes.}
    \label{tab:attribute_performance}
\end{table*}

\subsection{Experimental Setup}
\label{sec:experiments_setup}
All baselines (except Llama 3.1) are re-implemented using OPT-350M as a base-model. 
For the LLM prompting baseline, we query the instruction-tuned Llama 3.1-70B chat model~\citep{dubey2024llama} using a fixed two-message chat template: a system meta-instruction (similar to the meta-prompt used by MCTune~\citep{nguyen-etal-2024-multi}) that defines the available attribute tags (corresponding to Appendix~\ref{sec:index_list}) and their meanings, and a user message that lists the desired subset of $k$ targets as bracketed \texttt{[tag: value]} pairs; the model is instructed to output a JSON object with a single \texttt{text} field, and generations are capped at 100 tokens to match other baselines.
Moreover, inference-time algorithms use a linguistic discriminator (LD) to estimate the linguistic attributes of generated text. This component is independently pre-trained and frozen, allowing for differentiable computation of linguistic attributes and backpropagation of the error. It is trained on attributes generated by the explicit linguistic attribute extraction algorithms.

The \textbf{Linguistic Discriminator} (LD) is a crucial component for inference-time algorithms. It is pre-trained on data generated by the explicit linguistic attribute extraction algorithms (described in Section~\ref{sec:index_list}) to provide an efficient estimation of linguistic attributes for these methods.
It uses a DeBERTa encoder~\citep{he2021debertav3} with the token embedding layer replaced with that of OPT-350M, followed by a projection layer. The LD is trained to minimize the mean squared error between predicted attributes and gold attributes:
\[
\ell_{disc}(x) = \| \mathrm{LD}(x) - l^x \|_2^2.
\]
The final MSE loss of the pre-trained LD is $0.16$ on our test set. The correlation between the predicted MSE by the LD and the real MSE by the original linguistic attribute extractor tool is $0.8$, which is sufficiently high for reliable utilization.

We tune the hyperparameters of all baselines using grid search. The final hyperparameters used for each baseline are detailed in Appendix~\ref{app:baseline_hparams}. Further details on the training setup and computational infrastructure are provided in Appendix~\ref{app:experimental_details}.

\subsection{Datasets}
We use 6.8M samples (360M tokens) from publicly available datasets spanning multiple domains and writing styles, truncated to 100 tokens maximum. A detailed list is provided in Appendix~\ref{app:datasets}.

\subsection{Metrics}
Our model is evaluated on two key metrics.
Most Controlled Text Generation (CTG) papers use two primary metrics for evaluation: \textbf{attribute accuracy} and \textbf{fluency}.
Attribute accuracy measures how well the generated text adheres to the specified attributes, while fluency assesses the grammatical and logical coherence of the text.

\textbf{Mean Squared Error (MSE)} of attributes calculates the error between attributes of the generation and the desired target attributes.
With sentence length as an example, MSE measures the squared difference between the length of the generation and the target length.
Attribute measurements for MSE calculation are performed using the same explicit algorithms used to define the target attributes.
This provides an exact measure, making human evaluation for attribute accuracy unnecessary.
Generations may achieve a good score on the target attributes while being non-fluent (i.e., logically or grammatically incorrect).
To quantify the trade-off between control and fluency, we evaluate \textbf{Perplexity (PPL)} according to GPT2-XL~\citep{radford2019language}, following recent work such as~\citet{shen-huang-2025-llm} and~\citet{fathi2025unifying}.
To complement this automated metric, we also conduct human evaluations of fluency, detailed in Appendix~\ref{app:human_eval_details}.

\section{Results}

\subsection{Main Results}
\label{sec:main_results}

Table~\ref{tab:main_results} summarizes the performance of various models in controlled text generation tasks. The reported MSE is an average calculated across experiments controlling 1, 5, 10, 20, and 40 attributes simultaneously. Each attribute count setting was evaluated using 2,000 test samples per seed across 3 random seeds (6,000 generations per attribute count). For models evaluated across all five attribute counts, this totals 30,000 generations per model. Due to computational cost, the inference-time baselines (PPLM, Mix\&Match, COLD, PiLM) were only run for the 40-attribute setting (6,000 generations), and their MSE in Table~\ref{tab:main_results} corresponds to that setting. This design reflects overall performance across varying control complexities and mitigates potential bias from specific attribute selections or random seeds.

Vanilla OPT-350M, generating text without attribute control, provides a baseline with an average MSE of 2.00 and a perplexity of 12.7.
Models achieving an average MSE lower than 2.0 include LingGen, Mix\&Match, and PTG. Among those, Mix\&Match exhibits very high perplexity, indicating a sacrifice of fluency for control. LingGen achieves the best overall average MSE (1.08) while maintaining competitive perplexity (15.4). LingGen also demonstrates consistently low MSE across attribute counts and seeds (Table~\ref{tab:delta}), highlighting its scalability and robustness. In contrast, models like MCTune (OPT-350M), Guided Decoding, and Llama 3.1-70B show less stable performance as the number of attributes varies.

Furthermore, we conducted a human evaluation of text fluency for the top-performing fine-tuned models. As shown in Table~\ref{tab:human_eval}, LingGen achieves the highest average fluency score. Further details on the human evaluation setup and annotator guidelines are provided in Appendix~\ref{app:human_eval_details}. Qualitative samples in Appendix~\ref{sec:sample_comp} demonstrate the gap in existing models: PTG and Llama~3.1 either break attribute constraints or collapse to incoherent text, whereas LingGen maintains usable generations while achieving the target attributes.

To better understand model performance on specific types of attributes, Table~\ref{tab:attribute_performance} shows the Mean Absolute Error (MAE) for four representative attributes: sophisticated word count, total words, lexical words, and unique word ratio. LingGen achieves the lowest error across all these attributes.\looseness-1

The Llama 3.1-70B model, despite its strong general capabilities, shows higher average MSE (3.11) and middling performance on specific attributes (Table~\ref{tab:attribute_performance}), suggesting limitations in fine-grained control over linguistic attributes compared to specialized methods. Table~\ref{tab:deep_mae} further highlights this gap on deep structural attributes (syntactic: Complex Nominals, T-units; psycholinguistic: Age of Acquisition).

\begin{table}[t]
    \centering
    \small
    \begin{tabular}{lrrr}
    \toprule
    \textbf{Attribute} & \textbf{LingGen} & \textbf{Llama 70B} & \textbf{Gap} \\
    \midrule
    Complex Nominals & 0.84 & 2.36 & 2.8$\times$ \\
    T-units & 0.17 & 0.57 & 3.4$\times$ \\
    Age of Acquisition & 12.00 & 75.36 & 6.3$\times$ \\
    \bottomrule
    \end{tabular}
    \caption{Comparison of Mean Absolute Error (MAE) for LingGen and Llama 3.1-70B on deep structural attributes. Gap denotes the ratio between the errors.}
    \label{tab:deep_mae}
\end{table}

\begin{table*}[t]
    \centering
    \begin{tabular}{lrrrrrr}
    \toprule
    \textbf{Method} & \textbf{1} & \textbf{5} & \textbf{10} & \textbf{20} & \textbf{40} & \textbf{Average} \\
    \midrule
    Reference & - & - & - & - & 0.00 & 0.00 \\
    PPLM & - & - & - & - & 5.99 & 5.99 \\
    COLD & - & - & - & - & 3.99 & 3.99 \\
    Mix\&Match & - & - & - & - & 1.58 & 1.58 \\
    PiLM & - & - & - & - & 2.12 & 2.12 \\
    \midrule
    Vanilla OPT-350M & 1.96 & 1.99 & 2.00 & 2.02 & 2.03 & 2.00 $\pm$ 0.03 \\
    BOLT & 1.24 & 1.96 & 2.19 & 2.42 & 2.59 & 2.08 $\pm$ 0.47 \\
    Llama 3.1-70B & 4.22 & 3.22 & 2.63 & 2.36 & 3.11 & 3.11 $\pm$ 0.64 \\
    Guided Decoding & 5.56 & 2.21 & 3.83 & 3.87 & 3.24 & 3.74 $\pm$ 1.09 \\
    MCTune {\tiny OPT-350M} & 2.89 & 6.19 & 17.43 & 20.88 & 9.90 & 11.46 $\pm$ 6.75 \\
    MCTune {\tiny Llama 2-7B} & 3.15 & 2.39 & 2.06 & 2.07 & 2.51 & 2.44 $\pm$ 0.40 \\
    PTG & 2.11 & 1.96 & 2.29 & 1.61 & \textbf{0.60} & 1.71 $\pm$ 0.60 \\
    LingGen (ours) & \textbf{1.17} & \textbf{1.10} & \textbf{1.24} & \textbf{1.00} & 0.90 & \textbf{1.08} $\pm$ 0.12 \\
    \bottomrule
    \end{tabular}
    \caption{MSE for different models when controlling 1, 5, 10, 20, or 40 attributes simultaneously.}
    \label{tab:delta}
\end{table*}

Appendix~\ref{app:ablation_studies} includes ablation studies on P-MASKING's effects compared to other strategies, different base models, and attribute integration methods. Results indicate that LingGen with P-MASKING outperforms other masking strategies when evaluated on variable attribute numbers. Our ablation study on attribute integration methods (Table~\ref{tab:integration-ablation}) shows that injecting the attribute representation at the BOS token yields the best performance, lowering MSE by 36--65\% compared with alternative injection points (adding to all tokens, outputs, or logits) while maintaining perplexity comparable to the baseline.

\subsection{Robustness Across Attribute Counts}
\label{sec:delta_table}
To evaluate control effectiveness over varying attribute numbers, we compare the MSE of models when controlling 1, 5, 10, 20, or 40 attributes simultaneously (Table~\ref{tab:delta}). For each attribute count, 2,000 test samples were evaluated per random seed, and the experiment was repeated with three random seeds. This resulted in 6,000 evaluations per attribute count setting and 30,000 total evaluations per model for models evaluated across all five attribute counts. For a given attribute count, attributes were randomly selected for control from the full set of 40 for each seed. PPLM, Mix\&Match, COLD, and PiLM were only run for the 40-attribute setting due to computational cost. The values reported in Table~\ref{tab:delta} are the average MSE across the three seeds for that specific count. The ``Average'' column shows the mean MSE across all five attribute counts (with standard deviation across the five means).

\subsection{Analysis of Attribute Interaction Effects}

\begin{figure*}[t]
    \centering
    \includegraphics[width=0.8\linewidth]{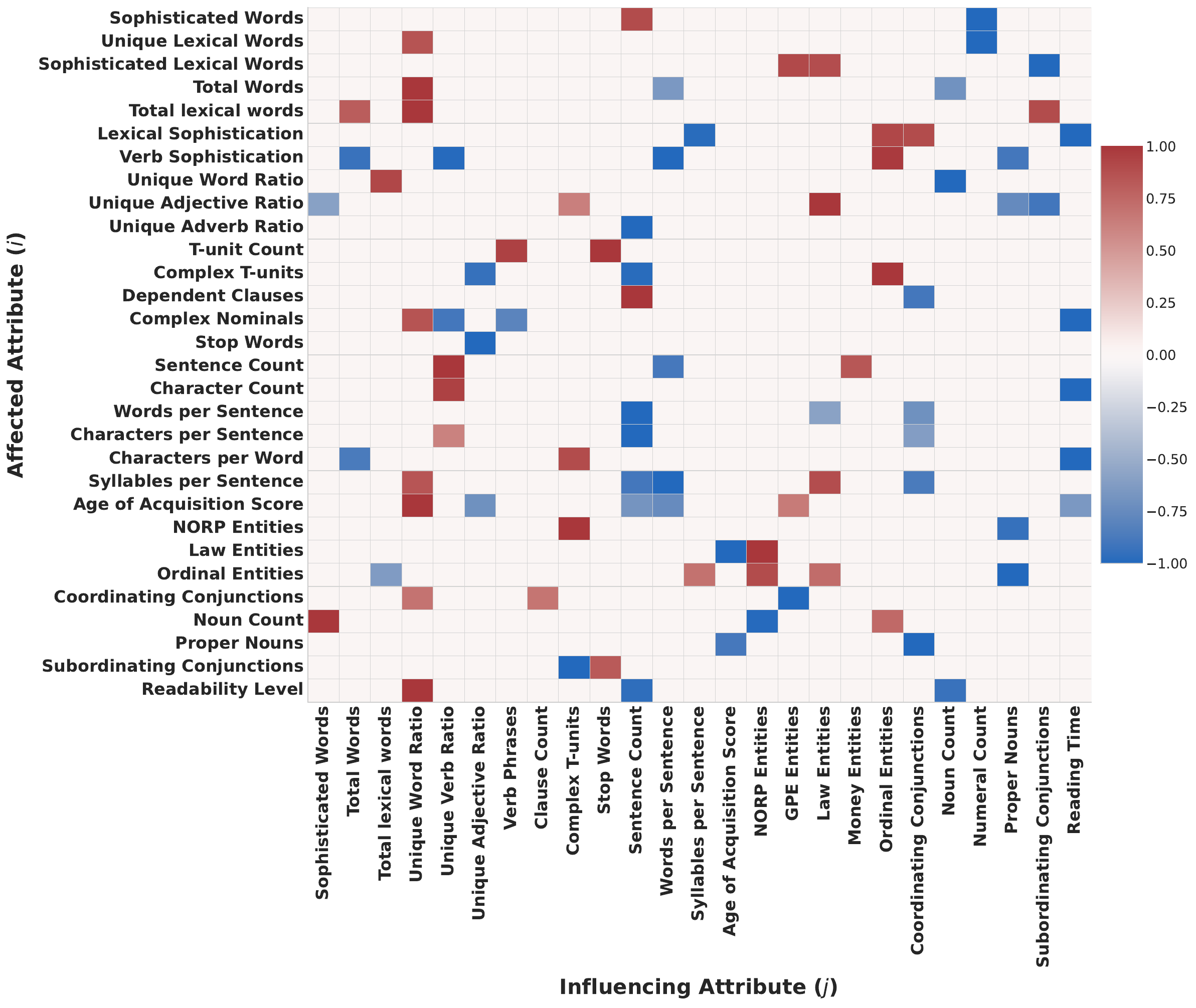}
    \caption{Interaction Effect ($\Delta \text{MSE}_{i \leftarrow j}$) of Controlling Attribute $j$ on Controlling Attribute $i$ (Statistically Significant Interactions, Row-Normalized). The cell values are the interaction effect ($\Delta \text{MSE}_{i \leftarrow j}$); Negative (blue) indicates synergy, while positive (red) indicates conflict.}
    \label{fig:interaction}
\end{figure*}

This experiment investigates the interaction effects between pairs of linguistic attributes in multi-attribute controlled text generation. Understanding these interactions is crucial, as controlling one attribute $j$ may facilitate (synergy) or hinder (conflict) the model's ability to control another attribute $i$. We quantify this interaction effect, $\Delta \text{MSE}_{i \leftarrow j}$, as the difference in the expected squared error for attribute $i$. Let $\se_i$ denote the squared error for attribute $i$. The interaction effect is then the difference in this value when attribute $j$ is included in the set of controlled attributes ($A_j$) versus when it is excluded ($\neg A_j$):
\[
    \Delta \mse_{i \leftarrow j} = \mathbb{E}[\se_i \mid A_j] - \mathbb{E}[\se_i \mid \neg A_j]
\]
A negative $\Delta \text{MSE}_{i \leftarrow j}$ indicates synergy (controlling $j$ improves control over $i$), while a positive value signifies conflict (controlling $j$ degrades control over $i$).

We conducted the analysis using the LingGen model. We evaluated 2,000 test samples per run across 8 random seeds (16,000 total samples). In each run, 20 attributes were randomly selected from the total 40 for simultaneous control. This setup ensures a balanced comparison, as any attribute $j$ had approximately a 50\% chance of being included in the controlled set. We report only statistically significant interactions (paired $t$-tests, $p < 0.05$). To facilitate comparison across different target attributes $i$, the effect sizes presented in Figure~\ref{fig:interaction} are row-normalized, scaling the strongest significant positive or negative effect for each attribute $i$ to $+1.0$ or $-1.0$, respectively.

Our analysis identifies attributes that have large effects when included as conditioning factors. {\em Sentence Count} shows synergies with metrics related to average sentence length, such as {\em Words per Sentence}, {\em Characters per Sentence}, and {\em Syllables per Sentence}, likely because fixing the sentence count provides a stable denominator for these averages. However, it conflicts with controlling {\em Dependent Clauses}, suggesting a tension between controlling sentence quantity and sentence complexity. Similarly, controlling {\em Words per Sentence} can help constrain related metrics such as total words and number of syllables.

Conversely, controlling for {\em Unique Word Ratio}, a measure of lexical diversity, consistently introduces conflicts. It degrades the ability to control {\em Total Words}, {\em Total lexical words}, {\em Age of Acquisition Score (AoA)}, and {\em Readability Level}. This suggests an inherent trade-off: enforcing high lexical diversity makes it harder for the model to adhere to specific length constraints (which might otherwise favor repetition) or to use simpler and more common vocabulary (lower AoA and higher readability).

Controlling for higher-level composite attributes could improve control over their constituent components. {\em Reading Time For Average Readers} shows strong synergistic effects, improving control over {\em Lexical Sophistication}, {\em Complex Nominals}, {\em Character Count}, and {\em Characters per Word}. This implies that targeting an overall reading time forces the model to implicitly manage underlying factors like word complexity, length, and structure, thereby aiding explicit control over these factors.\looseness-1

Moreover, strong synergies are observed between definitionally linked attributes such as {\em Subordinating Conjunctions} and {\em Complex T-units} likely because complex T-units often require subordinating conjunctions. We also observe an unexpected synergy where controlling {\em Numeral Count} improves control over {\em Sophisticated Words} and {\em Unique Lexical Words}. One hypothesis is that numeral constraints induce a shift toward a enumerations, measurements, and comparisons, which increase reliance on self-contained constructs like nouns and domain-specific words, which perhaps improve control over sophisticated/unique words.\looseness-1

Finally, attribute compatibility matters: some controls systematically synergize while others conflict, suggesting that some attribute combinations may be difficult to satisfy simultaneously. They also motivate controlling a broader set of attributes when targeting a complex style, since changing one attribute can induce measurable shifts in others.\looseness-1

\section{Conclusion}
We presented LingGen, a controlled text generation method for fine-grained linguistic control with a variable number of real-valued attributes. LingGen combines a dedicated attribute encoder with BOS-based integration and achieves the strongest overall control-fluency-efficiency trade-off among the evaluated baselines, including settings that control up to 40 attributes simultaneously. We also introduced P-MASKING, a Pareto-sampled attribute masking strategy that improves robustness across attribute subset sizes.
Beyond aggregate control error, our interaction analysis indicates that fine-grained linguistic control is affected by synergies and conflicts among attributes.
First, for composite targets, controlling a broader set of related attributes can stabilize control by anchoring shared components. Second, some attribute pairs are inherently difficult to satisfy simultaneously, and effective control may require relaxing or reweighting constraints.
Future work includes extending beyond a fixed attribute set (e.g., natural-language attribute descriptions) and interaction-aware constraint selection.


\newpage
\section*{Limitations}
Our study focuses on a fixed set of 40 attributes computed by deterministic extractors, and LingGen does not directly support controlling unseen attributes without defining (and validating) new extractors. We also evaluate generations truncated to 100 tokens; maintaining fine-grained control over longer documents may introduce additional challenges (e.g., drift in attribute satisfaction over time).

We note that this work does not address ``few-to-many'' generalization in the sense of training on a small fixed subset of attributes and generalizing to control unseen attributes at test time. Instead, our focus is on robustness across the combinatorial space of attribute subsets: given a fixed set of $k$ attributes, P-MASKING enables the model to robustly handle any subset of size 1 to $k$ at inference, without retraining. Table~\ref{tab:delta} demonstrates this capability, showing stable performance across varying numbers of simultaneously controlled attributes.

\section*{Ethical Statement}
Fine-grained control over linguistic attributes can be beneficial (e.g., readability adaptation) but may also enable misuse, such as producing persuasive or targeted text optimized for manipulation, impersonation, or misinformation. Because our controls include stylistic and complexity attributes, a malicious user could attempt to tailor text to specific audiences or contexts. Mitigations include restricting deployment to trusted settings, applying safety filters and content policies at generation time, and auditing attribute extractors and training data for biases that could be amplified through controlled generation.

\bibliography{custom,anthology_part1,anthology_part2}

\clearpage
\appendix

\section{Representative Sample Comparison}
\label{sec:sample_comp}

Table~\ref{tab:quantitative_comparison} shows generated texts for four sets of target attributes. 

\begin{table*}[t!]
    \centering
    \renewcommand{\arraystretch}{1.2} 
    \begin{tabularx}{\textwidth}{@{} l >{\RaggedRight}X >{\RaggedRight}X >{\RaggedRight}X @{}}
    \toprule
    \textbf{Sample Task} & \textbf{PTG (Baseline)} & \textbf{Llama 3.1 (Prompted)} & \textbf{LingGen (Ours)} \\
    \midrule
    
    \rowcolor{HeaderGray}
    \multicolumn{4}{p{\dimexpr\textwidth-2\tabcolsep\relax}}{\textbf{Task 1: Deep Control.} \Target{Sentences (4), Soph. Words (11), Complex Nominals (7)}} \\
    \textbf{Text} & 
    \textit{...and is a web-based application that allows you to manage your business... For example, Firefox, Chrome, Opera, Safari, Internet Explorer, Mozilla, etc.} & 
    \textit{The intricate complexities of modern society necessitate an elevated comprehension of nuanced dialectics... an examination of societal hierarchies reveals a penchant for grandiose rhetoric.} & 
    \textit{The new version of the "Sonic" is now available for download. This entry was posted on Friday, June 1st, 2018... Pinging is currently not allowed.} \\
    \textbf{Analysis} & 
    \Success\ \Attr{Sentences}{4}{4} \newline
    \Failure\ \Attr{Soph. Words}{11}{9} \newline
    \Failure\ \Attr{Complex Nom.}{7}{5} \newline
    \Note{Fluency Failure: Incoherent.} & 
    \Success\ \Attr{Sentences}{4}{4} \newline
    \Failure\ \Attr{Soph. Words}{11}{30} &
    \Success\ \Attr{Sentences}{4}{4} \newline
    \Failure\ \Attr{Soph. Words}{11}{9} \newline
    \Failure\ \Attr{Complex Nom.}{7}{4} \\
    \midrule
    
    \rowcolor{HeaderGray}
    \multicolumn{4}{p{\dimexpr\textwidth-2\tabcolsep\relax}}{\textbf{Task 2: Simple Lexical Constraint.} \Target{Soph. Word Count (0), Unique Word Ratio (1)}} \\
    \textbf{Text} & 
    \textit{The following is a list of all the articles on the website. This entry was posted in Uncategorized and tagged American History...} &
    \textit{cat the dog} &
    \textit{A man is sitting down on his bike.} \\
    \textbf{Analysis} &
    \Failure\ \Attr{Soph. Words}{0}{10} \newline
    \Failure\ \Attr{Unique Ratio}{1.0}{0.73} \newline
    \Note{Fluency Failure: Catastrophic collapse.} &
    \Success\ \Attr{Soph. Words}{0}{0} \newline
    \Success\ \Attr{Unique Ratio}{1.0}{1.0} &
    \Success\ \Attr{Soph. Words}{0}{0} \newline
    \Success\ \Attr{Unique Ratio}{1.0}{1.0} \newline
    \Note{Fluent and coherent.} \\
    \midrule
    
    \rowcolor{HeaderGray}
    \multicolumn{4}{p{\dimexpr\textwidth-2\tabcolsep\relax}}{\textbf{Task 3: Syntactic Structure.} \Target{Sentences (2), Dependent Clauses (3), Words/Sentence (12.5)}} \\
    \textbf{Text} & 
    \textit{This is a great way to get your kids involved in the arts and crafts! ... For ages 3-5.} &
    \textit{The sun was shining brightly in the clear blue sky. The birds were singing their sweet melodies from the trees.} &
    \textit{The first episode of the new series, The Vampire Diaries. This is the first episode of The Vampire Diaries.} \\
    \textbf{Analysis} &
    \Failure\ \Attr{Sentences}{2}{4} \newline
    \Failure\ \Attr{Dep. Clauses}{3}{0} &
    \Success\ \Attr{Sentences}{2}{2} \newline
    \Failure\ \Attr{Dep. Clauses}{3}{0} &
    \Success\ \Attr{Sentences}{2}{2} \newline
    \Failure\ \Attr{Dep. Clauses}{3}{0} \\
    \midrule
    
    \rowcolor{HeaderGray}
    \multicolumn{4}{p{\dimexpr\textwidth-2\tabcolsep\relax}}{\textbf{Task 4: Surface Constraints.} \Target{Sentences (1), Words (5), Characters (21)}} \\
    \textbf{Text} & 
    \textit{http://www.c-c.org/c/b/p/n/d/...} & 
    \textit{They are very happy now.} & 
    \textit{There are people on bikes.} \\
    \textbf{Analysis} & 
    \Failure\ \Attr{Sentences}{1}{0} \newline
    \Failure\ \Attr{Words}{5}{1} \newline
    \Failure\ \Attr{Characters}{21}{131} \newline
    \Note{Fluency Failure: Generated a URL.} & 
    \Success\ \Attr{Sentences}{1}{1} \newline
    \Failure\ \Attr{Words}{5}{4} \newline
    \Failure\ \Attr{Characters}{21}{22} & 
    \Success\ \Attr{Sentences}{1}{1} \newline
    \Success\ \Attr{Words}{5}{5} \newline
    \Failure\ \Attr{Characters}{21}{25} \\
    \bottomrule
    \end{tabularx}
    \caption{Quantitative analysis of model performance on four diverse generation tasks. Values are computed on the displayed excerpts. For discrete targets, \Success denotes an exact match; \Failure denotes a mismatch or a clear fluency failure. The examples illustrate both successes and failures across methods.}
    \label{tab:quantitative_comparison}
    \end{table*}

\section{Datasets Used}
\label{app:datasets}
The datasets used in our experiments span a range of domains and writing styles, including web text (C4), paraphrase pairs (MRPC), question pairs (QQP), and natural language inference datasets (ANLI, RTE, STS-B, SNLI, MNLI, FeverNLI). All datasets are utilized as single text samples, focusing on characteristics such as user-generated content, formally written text, automatically generated text, etc. The following datasets were used in our experiments: Common Crawl (C4)~\citep{raffel2020exploring}, Microsoft Research Paraphrase Corpus (MRPC)~\citep{dolan-brockett-2005-automatically}, Quora Question Pairs (QQP)~\citep{QQP}, Adversarial NLI (ANLI)~\citep{nie-etal-2020-adversarial}, Recognizing Textual Entailment (RTE)~\citep{dagan2005pascal}, Semantic Textual Similarity Benchmark (STS-B)~\citep{cer-etal-2017-semeval}, Stanford Natural Language Inference (SNLI)~\citep{bowman-etal-2015-large}, Multi-Genre Natural Language Inference (MNLI)~\citep{williams-etal-2018-broad}, and FeverNLI~\citep{thorne-etal-2018-fever}.

\section{List of Linguistic Attributes}
\label{sec:index_list}

We use expert-crafted linguistic complexity attributes as the control attributes for CTG.
For the full descriptions please refer to~\citet{Lu_2010},~\citet{lu2012relationship}, and~\citet{lee-lee-2023-lftk}.
Briefly, \textbf{Automated Readability Index} measures text complexity, \textbf{Lexical words} are content words (nouns, verbs, adjectives, adverbs), and \textbf{Sophisticated words} are less frequent words in the American National Corpus.
\textbf{Age of Acquisition} refers to the age at which a word is typically learned.
Table~\ref{tab:indices} lists all the attributes that we use.

\begin{table}[h!]
    \centering
    \begin{tabular}{ l }
    \toprule
    \# Unique sophisticated words\\
    \# Unique lexical words \\
    \# Unique sophisticated lexical words\\
    \# Total words\\
    \# Total sophisticated words\\
    Lexical sophistication (unique) \\
    Verb sophistication\\
    Ratio of unique words\\
    Ratio of unique verbs\\
    Ratio of unique adjectives\\
    Ratio of unique adverbs\\
    \# Dependent clauses \\
    \# Clauses \\
    \# T-units \\
    \# Complex T-units \\
    \# Complex nominals \\
    \# Stop Words\\
    \# Sentences\\
    \# Characters\\
    Average Words Per Sentence\\
    Average Characters Per Sentence\\
    Average Characters Per Word\\
    Average Syllables Per Sentence\\
    Total Age Of Acquisition Of Words\\
    \# Named Entities Norp\\
    \# Named Entities Gpe\\
    \# Named Entities Law\\
    \# Named Entities Money\\
    \# Named Entities Ordinal\\
    \# Coordinating Conjunctions\\
    \# Nouns\\
    \# Numerals\\
    \# Proper Nouns\\
    \# Subordinating Conjunctions\\
    Automated Readability Index\\
    Reading Time For Average Readers\\
    \bottomrule
    \end{tabular}
    \caption{Linguistic attributes used in this paper.}
    \label{tab:indices}
\end{table}

\section{Additional Experimental Details}
\subsection{Training Details}
\label{app:experimental_details}
LingGen is trained using LoRA with parameters $r=64$ and $\alpha=128$, and a batch size of 140, using the AdamW optimizer over 3 epochs. The model is selected based on the best validation step. We set a maximum sequence length of 100 tokens, and train on a single A100 40GB GPU.

We evaluate two variants of MCTune: one using \textbf{Llama 2-7B} as the base model, and another using \textbf{OPT-350M} and trained on the same dataset as LingGen.
Because MCTune requires In-Context Fine-Tuning (ICFT) with a long prompt, and a context up to 1024 tokens, it takes 216 GPU hours to train, while LingGen only takes 18 GPU hours.
Subsequently, MCTune was trained on an HPC using 12 GPUs.

\subsection{Baseline Hyperparameters}
\label{app:baseline_hparams}

We tune the hyperparameters of all baselines using grid search over common ranges suggested in their respective papers and report the best performing configuration found on our validation set. The key hyperparameters used in our experiments are shown in Table~\ref{tab:baseline_hparams}.

For Guided Decoding, we adapt the logit adjustment formula of FUDGE~\citep{yang-klein-2021-fudge} to our multi-attribute control task.
\begin{equation}
\log p + \lambda \times \varphi,
\end{equation}
where $\varphi$ is the favor signal.
Due to the varying difficulty and MSE scales across different attribute sets, we set
\begin{equation}
\varphi = 1/\text{MSE}
\end{equation}
(as lower MSE is better) and then normalize it per sample as
\begin{equation}
\varphi_{\text{norm}} = \frac{\varphi - \varphi_{\min}}{\varphi_{\max} - \varphi_{\min}}.
\end{equation}
This normalization makes the weighting parameter $\lambda$ less sensitive to the scale of individual attribute errors.

\begin{table}[h!]
    \centering
    \small
    \begin{tabular}{lll}
    \toprule
    \textbf{Method} & \textbf{Hyperparameter} & \textbf{Value} \\
    \midrule
    Guided Decoding & $\lambda$ & 5 \\
    \midrule
    \multirow{5}{*}{PPLM} & window\_length & 5 \\
    & grad\_length & 20 \\
    & stepsize & 0.01 \\
    & gamma & 1 \\
    & num\_iterations & 10 \\
    \midrule
    \multirow{4}{*}{COLD} & stepsize & 0.1 \\
    & constraint-weight & 0.5 \\
    & topk & 10 \\
    & num-iters & 2000 \\
    \midrule
    BOLT & learning\_rate & 0.05 \\
    \midrule
    \multirow{5}{*}{Mix\&Match} & max\_iter & 8 \\
    & n\_samples & 2 \\
    & alpha & 10 \\
    & beta & 1 \\
    & temperature & 1 \\
    \midrule
    \multirow{4}{*}{PiLM} & stepsize & 0.1 \\
    & M & 2 \\
    & future\_n\_tokens & 5 \\
    & ppl\_weight & -0.3 \\
    \midrule
    \multirow{3}{*}{MCTune} & batch\_size & 48 \\
    & lr & 2e-5 \\
    & warmup\_steps & 1000 \\
    \midrule
    \multirow{2}{*}{PTG} & batch\_size & 320 \\
    & lr & 5e-5 \\
    \bottomrule
    \end{tabular}
    \caption{Hyperparameters used for baseline methods.}
    \label{tab:baseline_hparams}
\end{table}

\section{Ablation Studies}
\label{app:ablation_studies}
To understand the contributions of our proposed P-MASKING strategy, the impact of different base models, and the effect of attribute integration methods, we conducted ablation studies.

\paragraph{Ablation Study: Impact of P-MASKING}
We evaluated various versions of our model using different methods of masking attributes during training. The methods included:
\begin{itemize}
    \item \textbf{LingGen (No Masking)}: Attributes are not masked during training, serving as a baseline to assess the impact of masking.
    \item \textbf{LingGen (Dropout)}: A fixed dropout rate of 0.3 is applied to the attributes, introducing randomness to the training process.
    \item \textbf{LingGen (Fixed Rate)}: A fixed masking rate of 0.3 is applied, providing a consistent level of attribute masking.
    \item \textbf{LingGen (P-MASKING)}: Our proposed P-MASKING strategy, which adapts the masking rate based on a power law distribution.
\end{itemize}

\begin{table}[h!]
    \centering
    \small
    \begin{tabular}{lcc}
    \toprule
    \textbf{Method} & \textbf{MSE $\downarrow$} & \textbf{PPL $\downarrow$} \\
    \midrule
    No Masking & 1.01 & 17.4 \\
    Fixed Rate & 1.13 & 16.4 \\
    P-MASKING & \textbf{0.90} & \textbf{16.3} \\
    \bottomrule
    \end{tabular}
    \caption{Comparison of masking strategies using OPT-350M base model. P-MASKING achieves the lowest MSE and perplexity among the masking variants.}
    \label{tab:masking-ablation}
\end{table}

Table~\ref{tab:masking-ablation} shows that P-MASKING achieves lower MSE (0.90) than No Masking (1.01) and Fixed Rate masking (1.13), while also maintaining the lowest perplexity among these variants (16.3).

\paragraph{Impact of Base Model}
We further evaluated LingGen with our proposed P-MASKING strategy using different base LLMs, specifically GPT-2~\citep{radford2019language} and Pythia-410M~\citep{biderman2023pythia}. 

\begin{figure*}[t]
    \centering
    \includegraphics[width=0.55\linewidth]{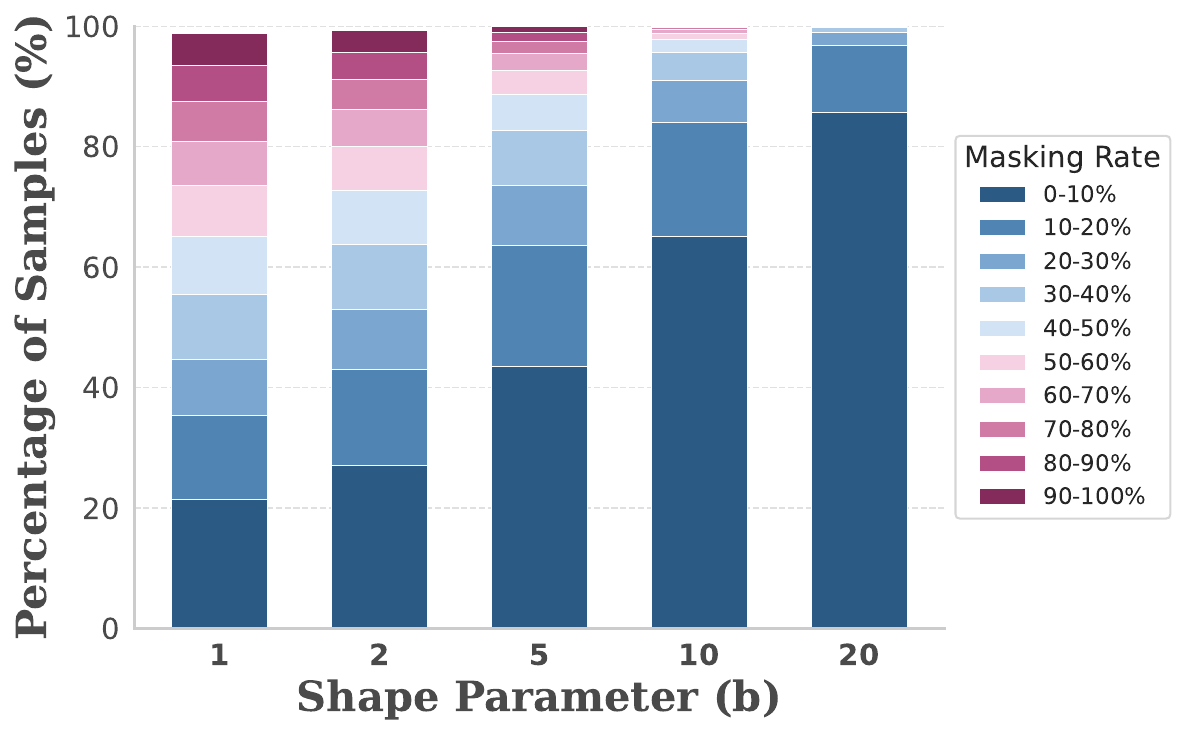}
    \caption{Effect of the Pareto shape parameter $b$ on the distribution of sampled masking proportions. Each bar aggregates Monte Carlo samples into masking-rate buckets, with darker segments indicating higher masking.}
    \label{fig:pareto_shape}
\end{figure*}

\begin{table}[h!]
    \centering
    \small
    \begin{tabular}{lcr}
    \toprule
    \textbf{Model} & \textbf{MSE $\downarrow$} & \textbf{PPL $\downarrow$} \\
    \midrule
    \textbf{Pythia-410M} & & \\
    \hspace{1em}No Masking & 2.58 & 14.4 \\
    \hspace{1em}Fixed Rate & 2.39 & 15.6 \\
    \hspace{1em}P-MASKING & \textbf{2.04} & \textbf{16.3} \\
    \midrule
    \textbf{GPT-2} & & \\
    \hspace{1em}No Masking & 2.69 & \textbf{19.8} \\
    \hspace{1em}Fixed Rate & 3.75 & 32.9 \\
    \hspace{1em}P-MASKING & \textbf{2.47} & 26.2 \\
    \bottomrule
    \end{tabular}
    \caption{Performance comparison of masking strategies across different base models (Pythia-410M, GPT-2). P-MASKING generally yields lower MSE compared to No Masking or Fixed Rate masking.}
    \label{tab:model-ablation}
\end{table}

As shown in Table~\ref{tab:model-ablation}, P-MASKING yields lower MSE than both No Masking and Fixed Rate masking for GPT-2 and Pythia-410M.

\paragraph{Impact of Different Integration Methods}
We also explored the effects of different methods for integrating linguistic attributes into the model. The integration methods compared were:
\paragraph{LingGen (Add to BOS):} Our proposed method, where the encoded attribute representation is added to the BOS token embedding.
\paragraph{LingGen (Add to All):} The encoded attribute representation is added to all decoder inputs at each time step.
\paragraph{LingGen (Add to Output):} The encoded attribute representation is added to the decoder output at each time step.
\paragraph{LingGen (Add to Logits):} The encoded attribute representation is added to the logits at each time step.
These methods were evaluated both with and without P-MASKING, using the OPT-350M base model. As shown in Table~\ref{tab:integration-ablation}, adding the encoded attribute representation to the BOS token embedding yields the lowest MSE in both settings. Notably, adding the attribute representation to all tokens produces extremely high perplexity (733--769), while BOS injection preserves competitive perplexity.

\begin{table}[h!]
    \centering
    \small
    \begin{tabular}{lcr}
    \toprule
    \textbf{Integration Variant} & \textbf{MSE $\downarrow$} & \textbf{PPL $\downarrow$} \\
    \midrule
    \textbf{No Masking} & & \\
    \hspace{1em}BOS (ours) & \textbf{1.01} & \textbf{17.4} \\
    \hspace{1em}All & 2.60 & 768.9 \\
    \hspace{1em}Output & 1.11 & 19.4 \\
    \hspace{1em}Logits & 1.52 & 19.8 \\
    \midrule
    \textbf{P-MASKING} & & \\
    \hspace{1em}BOS (ours) & \textbf{0.90} & \textbf{16.3} \\
    \hspace{1em}All & 3.52 & 733.8 \\
    \hspace{1em}Output & 1.76 & 16.3 \\
    \hspace{1em}Logits & 1.31 & 16.4 \\
    \bottomrule
    \end{tabular}
    \caption{Evaluation of attribute integration methods (using OPT-350M base model), with and without P-MASKING. Adding attribute information to the BOS token consistently provides the best MSE and competitive perplexity.}
    \label{tab:integration-ablation}
\end{table}

\section{Pareto Shape Parameter Study}
\label{app:pareto_shape}

Figure~\ref{fig:pareto_shape} compares candidate shape parameters by plotting the empirical masking-rate histograms (10k draws per setting). Smaller $b$ values allocate substantial mass to high masking rates, which reduces the model's exposure to dense attribute configurations and harms control accuracy. Conversely, very large $b$ values collapse the distribution toward minimal masking, curtailing the sparse cases needed for robustness. The selected $b=5$ concentrates a clear majority of samples in the $0$--$30\%$ bucket while leaving a visible tail across the higher buckets.

\section{Human Evaluation Details}
\label{app:human_eval_details}

To complement the automatic metrics (MSE and PPL), we conducted a human evaluation focusing on the fluency of the generated text.

We randomly selected 40 unique prompts (sets of target linguistic attributes) from our test set of 2,000 prompts. For each of these 40 prompts, we took the corresponding generated text output from the top-performing fine-tuned models identified in our experiments (BOLT, PiLM, PTG, and LingGen). Each of these generated text samples was independently rated by two human annotators, who are members of the institution conducting this research. Annotators were asked to evaluate the fluency of each text on a 5-point Likert scale, ranging from 1 (Not fluent at all) to 5 (Perfectly fluent). The average scores for each method are reported in Table~\ref{tab:human_eval}.
Cohen's Kappa was 0.32, which indicates fair agreement.

It is worth noting the potential discrepancy between automated perplexity (PPL) scores (Table~\ref{tab:main_results}) and human fluency judgments (Table~\ref{tab:human_eval}). While PPL measures the model's confidence in predicting the sequence based on a general language model (GPT2-XL), human evaluation captures aspects like grammatical correctness, coherence, and naturalness. For instance, a model might generate repetitive or simplistic text that achieves low PPL but is rated lower by humans, or conversely, generate more complex but slightly awkward sentences that humans find less fluent despite potentially reasonable PPL. In our results, LingGen achieves the highest human fluency score, even though some models like PPLM or BOLT report slightly lower PPL values, highlighting the value of human assessment alongside automatic metrics.

The annotators were provided with the following guidelines to ensure consistency in their evaluations:

\begin{tcolorbox}[
    enhanced,
    breakable,
    skin first=enhanced,
    skin middle=enhanced,
    skin last=enhanced,
    colback=blue!5!white,
    colframe=blue!75!black,
    title=Human Evaluation Guidelines: Fluency
    ]
\textbf{Task Overview:} You will evaluate texts based on their \textbf{fluency} (how easy and natural they are to read).

\textbf{Definition:} Fluency measures whether a text reads smoothly and naturally, considering both grammatical correctness and logical flow. A fluent text should be effortless to process, with no structural awkwardness or confusing word choices.

\textbf{Rating Scale:}

\textbf{Score 5 (Perfectly fluent):} No grammatical errors, completely natural and effortless to read.

\hspace{1em}\textit{Example: ``After finishing her work, she decided to take a walk in the park to enjoy the beautiful weather.''}

\textbf{Score 4 (Very fluent):} Minor imperfections that barely affect readability (e.g., slightly awkward phrasing, but no grammatical errors).

\hspace{1em}\textit{Example: ``Having completed her work, she took a walk in the park for enjoying the weather.''}

\textbf{Score 3 (Moderately fluent):} Noticeable issues but main meaning is clear. May have minor grammatical errors or somewhat awkward phrasing that requires brief re-reading.

\hspace{1em}\textit{Example: ``After she finish her work, she decide to walk in the park because nice weather.''}

\textbf{Score 2 (Slightly fluent):} Multiple grammatical errors or incoherent structure. Requires significant effort to understand, but meaning can be extracted.

\hspace{1em}\textit{Example: ``She finishing work and walk park. Weather it was nice for her.''}

\textbf{Score 1 (Not fluent at all):} Severely broken grammar or completely incoherent. Nearly impossible to understand the intended meaning.

\hspace{1em}\textit{Example: ``Work finish she park walking nice it weather the.''}

\textbf{Decision Rules:}
\begin{itemize}
    \item \textbf{When in doubt:} Consider how much effort is required to understand the text. Choose a lower score if you had to re-read or mentally correct the text.
    \item \textbf{Grammar vs.\ coherence trade-off:} If grammar is perfect but logic is flawed (or vice versa), cap the score at 3. Both dimensions must be strong for scores of 4 or 5.
    \item \textbf{Context-specific:} Consider whether phrasing is natural for the specific linguistic context being evaluated (e.g., language learning level, domain-specific writing).
\end{itemize}

\textbf{Important Notes:}
\begin{itemize}
    \item \textbf{Do not} consider factual correctness, content quality, or completeness. Only focus on linguistic fluency.
    \item \textbf{Do:} Use the anchor examples as reference points throughout your evaluation.
\end{itemize}
\end{tcolorbox}

\end{document}